# Mitigating Asymmetric Nonlinear Weight Update Effects in Hardware Neural Network based on Analog Resistive Synapse

Chih-Cheng Chang, Pin-Chun Chen, Teyuh Chou, I-Ting Wang, Boris Hudec, Che-Chia Chang, Chia-Ming Tsai, Tian-Sheuan Chang, *Senior Member, IEEE*, and Tuo-Hung Hou, *Senior Member, IEEE*

*Abstract*—Asymmetric nonlinear weight update is considered as one of the major obstacles for realizing hardware neural networks based on analog resistive synapses because it significantly compromises the online training capability. This paper provides new solutions to this critical issue through co-optimization with the hardware-applicable deep-learning algorithms. New insights on engineering activation functions and a threshold weight update scheme effectively suppress the undesirable training noise induced by inaccurate weight update. We successfully trained a two-layer perceptron network online and improved the classification accuracy of MNIST handwritten digit dataset to 87.8/94.8% by using 6-bit/8-bit analog synapses, respectively, with extremely high asymmetric nonlinearity.

## I. Introduction

DEEP learning on software-defined neural networks has greatly advanced many aspects of artificial intelligence, such as image recognition, speech recognition, natural language understanding, and decision making, by using a general-purpose learning procedure on conventional computing hardware, such as central processing units or graphic processing units [1]. Recently, the limitations of the conventional computing hardware on the form factor, cost, and power consumption of deep learning systems have promoted active research on hardware neural networks (HNNs) that are inspired by the physical structure of high-density, low-power, and distributed biological neural networks [2-3]. The HNNs employing crossbar resistive random access memory (RRAM)-based synaptic arrays are particularly interesting because they may significantly improve the learning efficiency by fully parallelizing the vector-matrix multiplication (weighted sum) and outer-product weight update through the distributed weight storage [4]. Impressive RRAM-based HNN prototypes have been demonstrated including speech and electroencephalography signal recognition [5], pattern recognition [6], dot product engine [7], nature image processing [8], and face classification [9]. However, most implementations were limited to primitive algorithms such as single-layer perceptron or sparse coding, and many of them relied on offline training. By contrast, our brain processes information intelligently and in real time through a considerably more complicated deep neural networks with a layered structure [10]. The software approach also resorts to extremely deep neural networks to improve recognition accuracy that surpasses humans [11]. Designing a deep HNN with online training capability is a nontrivial task because of the challenges of non-ideal properties of RRAM synapses [4, 12–16]. In particular, the asymmetric nonlinear weight update of RRAM synapses had been identified as an outstanding issue that significantly degraded the learning accuracy [15–17]. Several approaches have been proposed aiming at improving the asymmetric nonlinear weight update of RRAM synapses, for example by careful material/device engineering [18], by implementing an additional compensational circuit [19], or by optimizing the operating pulse waveform [20]. However, to further increase the immunity against this undesirable nonlinear effect, co-optimization considering both the device characteristics and the HNN algorithms suitable for future deep learning would be essential.

This paper focused on the HNN design employing a Ta/HfO$_2$/Al-doped TiO$_2$/TiN synaptic device deposited by atomic layer deposition (ALD). The device exhibited bidirectional analog weight change that is suitable for the ultimate high-density neural network by using one two-terminal RRAM cell per synapse [4]. By contrast, more than one RRAM cell per synapse would be required by using binary RRAM [21] or unidirectional analog RRAM [15]. Furthermore, we presented a hardware-applicable multilayer perceptron (MLP) algorithm that accelerated supervised online training through massive parallelism. The MLP algorithm is the foundation of numerous deep-learning neural networks such as convolutional neural network (CNN). We reported that engineering the nonlinear activation function and employing a cutoff threshold on weight update were effective for mitigating the effects of asymmetric nonlinear weight update. The

This work was supported by the Ministry of Science and Technology of Taiwan under grant: MOST 105-2119-M-009-009 and Research of Excellence program MOST 106-2633-E-009-001, and Taiwan Semiconductor Manufacturing Company. T.-H. Hou acknowledges support by NCTU-UCB I-RiCE program, under grant MOST 106-2911-I-009-301.

The authors are with Department of Electronics Engineering and Institute of Electronics, National Chiao Tung University, Hsinchu, Taiwan (email:thhou@mail.nctu.edu.tw)

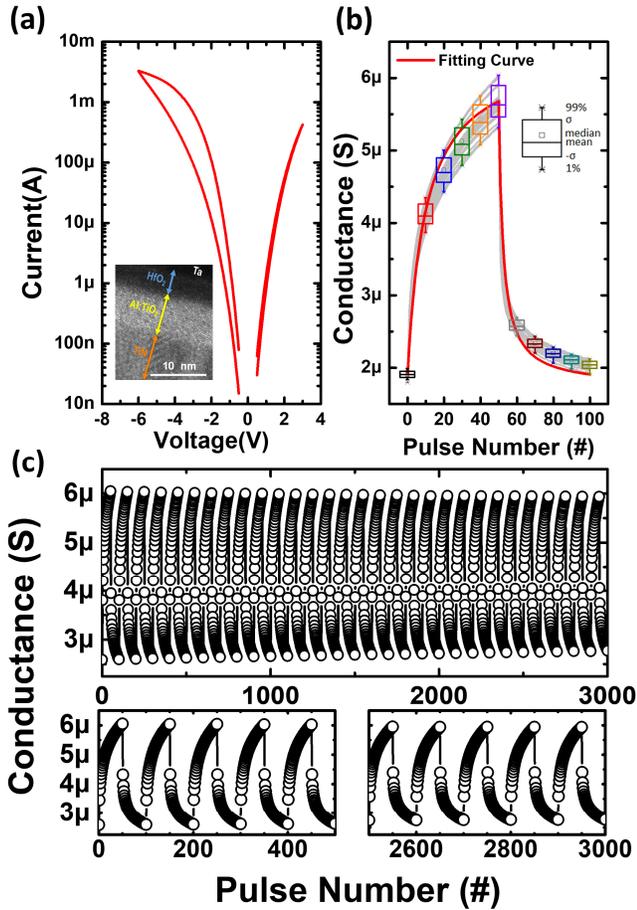

Fig. 1. (a) DC switching curve of the Ta/HfO$_2$/Al-doped TiO$_2$/TiN RRAM devices. Inset shows the cross-sectional TEM image of the Ta/HfO$_2$/Al-doped TiO$_2$/TiN RRAM device with a HfO$_2$/TiO$_2$ bilayer thickness of 12 nm. (b) Conductance modulation of multiple devices during AC potentiation and depression operations. Potentiation and depression pulses were 3 V for 0.01 ms and −3 V for 70 ms, respectively. The conductance value were readout at −1.5 V. The box-and-whisker plot indicating the device-to-device variation and the fitting curve using Eq. (1)-(3) are also sketched. (c) Reproducible conductance modulation during repeated potentiation and depression cycles with a total of 3,000 training pulses. The curve in the first and final 500 cycles are zoomed in.

classification accuracy of MNIST handwritten digit dataset was improved to 94.8% by considering an extremely high asymmetric nonlinearity in weight update. We also discussed the effects of reduced precision in weights and neurons. The results suggest the HNN algorithm, if designed properly, could be highly robust against the non-ideal asymmetric nonlinearity of weight update.

## II. FULLY ANALOG RRAM SYNAPSE

The synaptic device we considered in this study was implemented using a non-filamentary Ta/HfO$_2$/Al-doped TiO$_2$/TiN RRAM. The device resembled similar analog synaptic properties as those in the previously reported Ta/TaO$_x$/TiO$_2$/Ti RRAM [20, 22–23]. Figure 1(a) shows the DC switching characteristics of the device with positive set and negative reset voltages. The total film thickness of the new HfO$_2$/TiO$_2$ oxide bilayers, as shown in the transmission electron microscope image (Inset in Fig. 1(a)), was further scaled down to approximately 12 nm by using ALD compared with the previous 40-nm-thick TaO$_x$/TiO$_2$ deposited by sputtering. The conformal ALD growth and oxide thickness scaling are highly desirable in future 3D crossbar synaptic arrays with extremely high-density and high-connectivity [20]. In 3D crossbar synaptic arrays, which resemble the 3D neural network in the human brain [24], the oxide bilayers must be conformally deposited on the vertical sidewalls of deep trenches or vias. The pitch size of the trenches or vias and thus the ultimate packing density are determined by the thickness of the oxide bilayers. More detailed discussions on the device fabrication process and RRAM switching characteristics will be presented elsewhere. Figure 1(b) shows the synaptic plasticity determined by measuring the device conductance (G) at −1.5 V. The G value changed from a G minimum ($G_{min}$) to a G maximum ($G_{max}$) in a range of approximately 4 μS within one cycle of synaptic operation, which comprised 50 consecutive potentiation pulses (P-pulses) at 3 V for 0.01 ms, followed by another 50 consecutive depression pulses (D-pulses) at −3 V for 70 ms. Here the increment and decrement of the device conductance are referred to as potentiation and depression, respectively. The device-to-device variation was considerably small because of the uniform ALD growth. The multiple analog states were precisely trained according to P/D-pulse inputs provided by external neural circuits. The bidirectional analog weight modulation with minimal conductance fluctuation obtained in this device is inherently different from the conventional filamentary synapses [25]. Furthermore, Fig. 1(c) shows reproducible synaptic plasticity during repeated potentiation and depression cycles with a total of 3,000 training pulses.

Although this device demonstrated promising analog synaptic responses, several non-ideal properties compared with the weight presented by floating-point numbers in software-defined neural networks are noticeable, as summarized in Fig. 2. Their influence should be carefully evaluated when implementing HNNs. First, the precision of synaptic weight is limited because the total plasticity range $G_{max}/G_{min}$ is finite and precise control on the weight value is governed by the integer number of training pulses. The discrete weight values in this device are bounded between $G_{max}$ and $G_{min}$ with approximately a 5 to 6-bit precision. Second, the weight change is not linearly proportional to the P/D-pulse inputs. The change of device conductance is more dramatic during the first few P/D-pulses and saturates as the number of pulses increases, which is referred to as nonlinear weight update. More importantly, the potentiation and depression characteristics are not symmetric. The weight change induced by the P-pulses is much stronger at the low-conductance states while that induced by the D-pulses is much stronger at the high-conductance states, which is referred to as asymmetric weight update. This asymmetric nonlinear weight update creates a hysteresis loop [4] in Fig. 2 when using the number of P/D-pulses to update weights, and results in a considerable error. For example, the weight value undergoing 25 P-pulses is not identical as that undergoing 50 P-pulses plus 25 D-pulses. The asymmetric nonlinear relation between discrete G values and the cumulative number of P/D-pulses ($n_P/n_D$) is fitted mathematically by:

$$G_P(n_P) = G_{min} + A \times \frac{n_P}{n_P + e^k} \quad (1)$$

$$G_D(n_D) = G_{max} - A \times \frac{(N - n_D)}{(N - n_D) + e^k} \quad (2)$$

$$A = (G_{max} - G_{min}) \times (1 + \frac{e^k}{N}) \quad (3)$$

where $G_P$ and $G_D$ are functions describing conductance of potentiation and depression curves, respectively. $N$ is the maximum allowed number of input pulses, which defines the weight precision of RRAM devices, and $k$ is a fitting parameter. Furthermore, this non-ideal effect could be quantitatively defined using an asymmetric nonlinearity factor (ANL) as:

$$ANL = \left[ G_P(\frac{N}{2}) - G_D(\frac{N}{2}) \right] / (G_{max} - G_{min}) \quad (4)$$

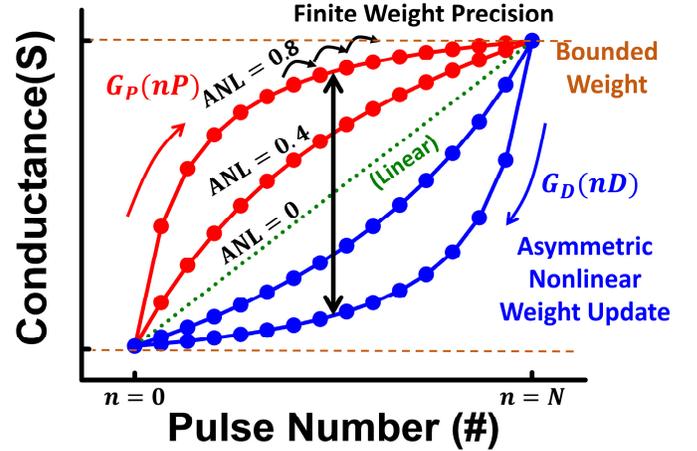

Fig. 2. Non-ideal properties of RRAM-based synapse including bounded weight with finite precision and asymmetric nonlinear weight update. The potentiation and depression curves with different degrees of asymmetric nonlinearity (ANL) are also plotted.

The ANL value is between zero to one. Completely symmetric potentiation and depression curves yield an ANL value of zero while the device exhibits an ANL value of 0.7 in Fig. 1(b).

III. MULTILAYER PERCEPTRON ALGORITHM FOR HNNs

Here we present a modified two-layer perceptron algorithm for implementing HNNs. The network includes one input neuron layer where neurons are numbered from 1 to $i$, one hidden neuron layer where neurons are numbered from 1 to $j$, and one output neuron layer where neurons are numbered from 1 to $k$, as illustrated in Fig. 3(a). Two adjacent neuron layers are referred to as pre-neuron and post-neuron layers along the forward propagation direction, and are fully connected using RRAM synapses. The fully connected network could be implemented using a crossbar RRAM array, as illustrated in Fig.

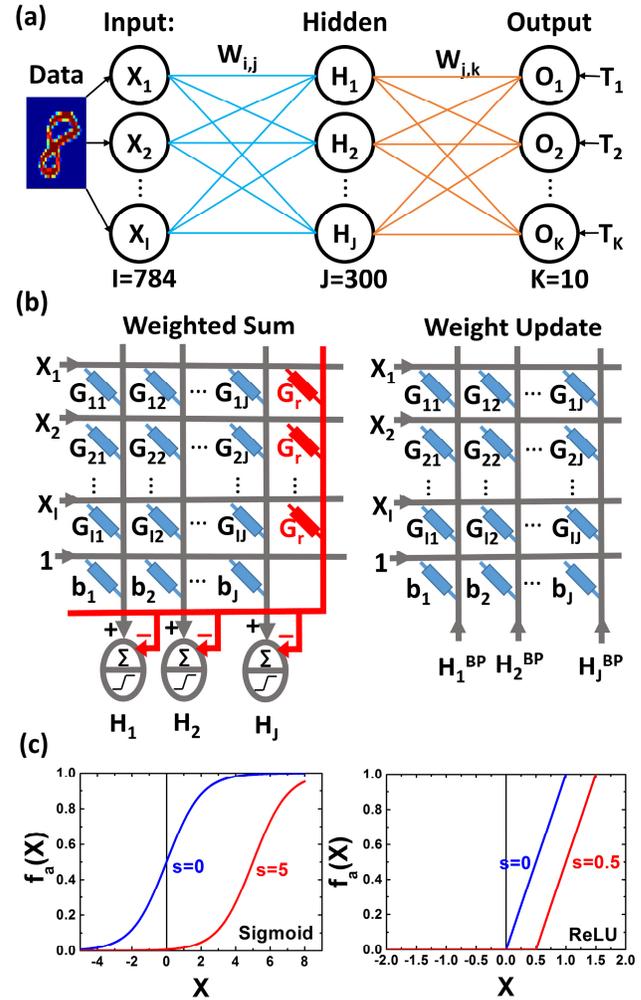

Fig. 3. (a) Illustration of a two-layer fully-connected perceptron network. (b) Implementation of weighted sum and outer-product weight update by using the crossbar RRAM configuration. Only the first fully-connected layer was illustrated. (c) Illustrations of sigmoid and ReLU activation functions with and without parallel shifts.

3(b). The initial conductance values of RRAM synapses are randomized in the range between $G_{min}$ and $G_{max}$. The weighted sum is performed by summing all current components on the same column, and the post neurons in different columns parallel convert analog current signal to digital voltage signal by using appropriate integrated circuits, such as a leaky integrate-and-fire neuron circuit. Note that time-consuming readout of conductance values of individual devices is not necessary when performing this weighted sum calculation by using the crossbar configuration. However, the synaptic weight values could be both positive and negative float-point numbers in conventional multilayer perceptron algorithms while the discrete conductance values in RRAM synapses are only positive. Therefore, the effect of signed weight values are emulated by subtracting a reference current flowing through a separate dummy column. The weighted sum values of the hidden ($H_j$) and output ($O_k$) neurons are given by:

$$H_j = f_a(S_j^H) = f_a \left[ b_j + \sum_{i=1}^{I} (X_i \times W_{ij}) \right] \quad (5)$$

$$= f_a \left\{ b_j + \sum_{i=1}^{I} \left[ X_i \times (G_{ij} - G_r) \right] \right\}$$

$$O_k = f_a(S_k^O) = f_a\left[b_k + \sum_{j=1}^{J}\left(H_j \times W_{jk}\right)\right] \quad (6)$$

$$= f_a\left\{b_k + \sum_{i=1}^{I}\left[H_j \times \left(G_{jk} - G_r\right)\right]\right\}$$

where $f_a$ is a nonlinear activation function, and $X_i$ is the value of input neurons. $S_j^H$ and $S_k^O$ are the weighted sum before activation. $b_j(b_k)$, $W_{ij}(W_{jk})$, and $G_{ij}(G_{jk})$ are bias value, signed weight value, and device conductance value in the first (second) fully-connected layer. $G_r$ is a constant reference conductance in the dummy column given by:

$$G_r = (G_{max} + G_{min})/2 \quad (7)$$

The nonlinear activation functions evaluated in this work include the popular sigmoid function and rectified linear unit (ReLU) function [26]. Fig. 3(c) shows the illustration of both functions. The sigmoid function is given by:

$$f_a(x) = \frac{1}{1+e^{-(x-s)}} \quad (8)$$

The ReLU function is given by:

$$f_a(x) = \max\left[0, (x-s)\right] \quad (9)$$

where $s$ is the parallel shifted value of the functions, which is set to zero by default. The supervised training is performed using a gradient-descent backpropagation method. The backpropagation values of the output ($O_k^{BP}$) and hidden ($H_j^{BP}$) neurons are given by:

$$O_k^{BP} = \left(T_k - O_k\right)f_a'\left(S_k^O\right) \quad (10)$$

$$H_j^{BP} = \sum_{k=1}^{K}\left(O_k^{BP}W_{jk}\right)f_a'\left(S_j^H\right) \quad (11)$$

where $T_k$ is the target value of each output neuron, and $f_a'$ is the derivative of the activation function. The conductance values of RRAM synapses are updated according to multiplications of forward-propagation values of pre-neurons and backward-propagation values of post-neurons, which is mathematically equivalent to perform outer product of two vectors. The multiplication could be realized through the careful overlapping of the pre- and post-neuron waveforms by using the crossbar configuration [16, 27], and the effective pulse number $\Delta n_{ij}$ ($\Delta n_{jk}$) applied on RRAM synapses in the first (second) fully-connected layer is given by:

$$\Delta n_{ij} = round(X_i \times \eta H_j^{BP}) \quad (12)$$

$$\Delta n_{jk} = round(H_j \times \eta O_k^{BP}) \quad (13)$$

where $\eta$ is the learning rate that is a scaling factor to regulate training. Rounding is performed considering only an integer of the effective pulse number. Positive and negative $\Delta n$ induce potentiation and depression of the device conductance, respectively. This outer-product weight update is highly efficient because all synapses in the array are updated simultaneously. More importantly, the update is solely based on the values of pre-neuron and post-neuron. No prior state of RRAM synapses needs be stored or readout before updating. If the weight change $\Delta W_{ij}(\Delta W_{jk})$ of RRAM synapses is linearly proportional to $\Delta n_{ij}$ ($\Delta n_{jk}$), Eqs. (12) and (13) would guarantee correct weight changes that follow the ideal gradient-descent backpropagation method. However, practical RRAM synapses possess nonlinear weight update behaviors as discussed in Eqs. (1) and (2). The new conductance value of the updated RRAM synapse, $G^{new}$, should be considered using Eqs. (1) and (2) as:

$$G^{new} = G_P(n_P^{old} + \Delta n) \quad \text{for } \Delta n > 0 \quad (14)$$
$$\quad\quad\ = G_D(n_D^{old} + \Delta n) \quad \text{for } \Delta n < 0$$

where $n_P^{old}$ and $n_D^{old}$ are equivalent cumulative number of P/D-pulses before update. Furthermore, rounding is performed on all forward- and backward-propagation values, i.e. $X_i$, $H_j$, $O_k$, $H_j^{BP}$, and $O_k^{BP}$, because the precision of practical neuron circuits is finite. For the mathematically unbounded activations, such as ReLU, an upper bound of neuron values was set to facilitate rounding in hardware. This upper bound was estimated in advance in this work to prevent clipping of high positive neuron values, but it potentially could be optimized empirically based on the classification results. This hardware-applicable algorithm could be extended to more sophisticate deep neural networks, such as CNN, with more than one hidden layer.

To evaluate the non-ideal properties of the RRAM analog synapse on the hardware-applicable MLP algorithm, we performed simulations by using MNIST handwritten digit dataset. The network size was fixed at 784×300×10. The bias values were set to zero for simplicity in this work. Each training epoch contained 60,000 training images, which were 28×28 pixels large with 256 gray levels. The weight update could be performed after every training image by using stochastic gradient descent or after a minibatch of training images by using minibatch gradient descent. We verified that the convergence errors of training did not yield significant differences between these two methods. Therefore, the results reported in this paper were obtained using minibatch gradient descent to accelerate the simulation. The typical batch size was set to 10 to 50. A constant learning rate was used throughout training, but the value was optimized individually for different hardware setting conditions to improve training convergence. The classification accuracy was tested after training by using another independent 10,000 testing images.

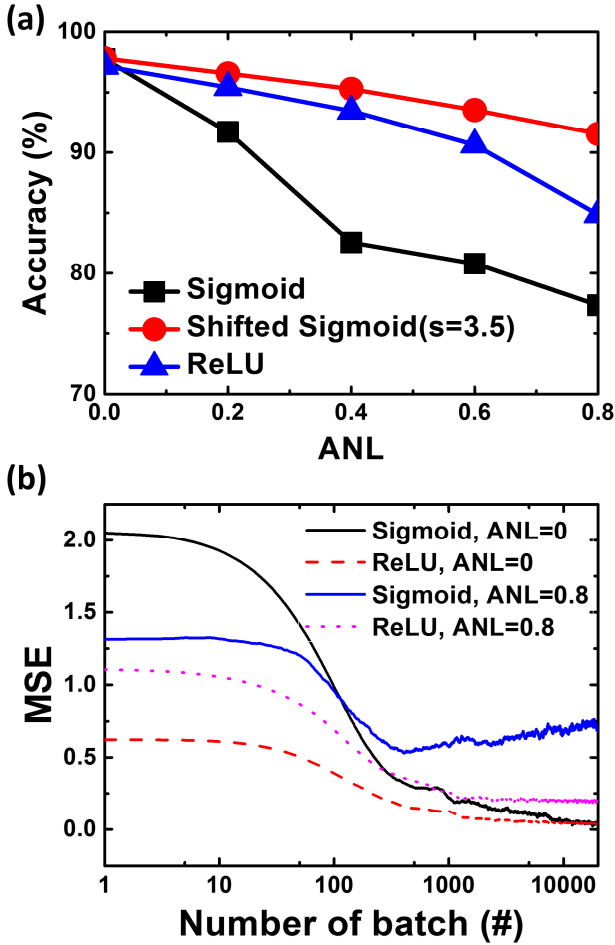

Fig. 4. Effects of various activation functions. (a) Dependence of asymmetric nonlinearity on the classification accuracy of MNIST handwritten digit dataset. (b) Convergence of MSE during training for both sigmoid and ReLU activation functions when linear and asymmetric nonlinear weight updates are assumed.

TABLE I
AVERAGE SPARSITY AND CLASSIFICATION ACCURACY BY USING SIGMOID AND RELU ACTIVATIONS WITH DIFFERENT SHIFTED VALUES $s$ AND ANL.

| Sigmoid | ANL=0 | | ANL=0.8 | |
|---|---|---|---|---|
| | Sparsity(%) | Accuracy (%) | Sparsity(%) | Accuracy (%) |
| s=0 | 36.06 | 97.92 | 0.12 | 77.33 |
| s=2 | 28.36 | 97.61 | 0.68 | 87.06 |
| s=2.5 | 27.72 | 97.63 | 3.24 | 89.59 |
| s=3 | 29.74 | 97.92 | 6.47 | 90.46 |
| s=3.5 | 32.69 | 97.91 | 11.32 | 91.55 |
| s=4 | 34.06 | 97.9 | 16.4 | 91.18 |
| s=4.5 | 37.85 | 97.93 | 23.79 | 89.86 |
| s=5 | 37.36 | 97.64 | 32.77 | 86.44 |
| ReLU | ANL=0 | | ANL=0.8 | |
| | Sparsity(%) | Accuracy (%) | Sparsity | Accuracy (%) |
| s=-0.1 | 53.78 | 96.72 | 40.71 | 84.28 |
| s=0 | 62.69 | 97.17 | 49.08 | 84.91 |
| s=0.1 | 65.87 | 97.42 | 55.15 | 85.19 |
| s=0.2 | 69.26 | 97.45 | 60.29 | 85.37 |
| s=0.3 | 74.52 | 97.32 | 65.2 | 84.88 |
| s=0.4 | 77.44 | 97.12 | 68.9 | 84.82 |

## IV. MITIGATING ASYMMETRIC NONLINEAR WEIGHT UPDATE BY ENGINEERING ACTIVATIONS

We first verified the effects of asymmetric nonlinear weight update on MLP training, which was reported to be one of the major limitations of electronic synaptic devices [15–17]. Both neuron and weight precision were set to 8 bits. The nonlinear but symmetric weight update did not significantly affect the classification accuracy (not shown), which is consistent with previous studies [15–17]. By contrast, Fig. 4(a) shows that the asymmetric nonlinear weight update substantially degraded the accuracy by using the sigmoid activation function and a practical ANL factor larger than 0.4. Figure 4(b) shows the convergence of mean square error (MSE) during training. The asymmetric nonlinear weight update resulted in a slower convergence rate and premature saturation of MSE compared with the linear one because the intended amount of weight change on the device is not linearly proportional to the effective pulse number according to Eqs (1)-(2). The amount of weight change depends on the current weight state and is not equal when receiving an identical number of P-pulses and D-pulses. Ideally corrections on Eqs (1)-(2) would be possible in the neuron circuits to compensate the effects of asymmetric nonlinear weight update if the current weight state is known. However, this would involve reading the weight (conductance) value of every individual RRAM synapse and significantly comprise the advantage of hardware acceleration through the crossbar configuration.

By contrast, we found that this undesirable effects could be effectively mitigated by using the ReLU activation. Both the classification accuracy in Fig. 4(a) and the convergence of training MSE in Fig. 4(b) were significantly improved when with a large ANL value of 0.8. Although both the ReLU and sigmoid functions rectify negative neuron values, the ReLU function produces a hard saturation effect where all negative values and their corresponding derivatives are forced to be exactly zero, i.e. both the values of $H_j$ and $H_j^{BP}$ or $O_k$ and $O_k^{BP}$ are zero. We referred those neurons with zero/non-zero values as inactive/active neurons. According to Eqs. (12)-(13), inactive neurons prevent those weights connected to them from updating ($\Delta n$=0). We define an average sparsity of the hidden layer as the fraction of zeros in $H_j$ [28]. For ANL=0.8, the ReLU function greatly increased the average sparsity to 49% compared with 0.12% when using the sigmoid function. The quasi-hard saturation effect could be employed by using shifted sigmoid activation functions, which also showed excellent immunity against asymmetric nonlinear update in Fig. 4(a). Table 1 list the average sparsity and classification accuracy by using the sigmoid and ReLU activation functions with different $s$ values for both linear (ANL=0) and highly nonlinear (ANL=0.8) weight update. The average sparsity increased with the increase of positive $s$, and the classification accuracy also increased with $s$ before a turning point. For ANL=0, both sigmoid and ReLU activations achieved high accuracy of over 97%. For ANL=0.8, the highest classification accuracy of 91.5% was obtained by using $s = 3.5$ for sigmoid while the highest classification accuracy of 85.4 % was obtained by using $s = 0.2$ for ReLU. Figure 5 illustrates that the optimized degrees of sparsity depended strongly on the choice of activation functions and the ANL values. The advantages of sparsity has been widely recognized in the field of deep neural networks [28]. The mathematical reasons behind had been explained by improving information disentangling, variable-size representation, linear separability, and distributed representation [28, 29]. In our MLP simulation, these factors were not determinant because the classification accuracy was relatively independent of the sparsity for the ideal linear update case. Both sigmoid and ReLU activations could achieve similar accuracy with different sparsity levels (~35% for sigmoid and ~65% for ReLU). We attribute the improvement by increasing the average sparsity for the highly nonlinear update case to the suppression of inaccurate weight update. The frequent switching of the signs of the small $\Delta n$ values during training would induce substantial update error and unnecessary perturbation toward convergence because of the asymmetric nonlinear weight update. Their effects could be considered as noise during weight training. The hard or quasi-hard saturation effect introduced by ReLU or shifting promotes convergence by suppressing noise induced by small $\Delta n$ while favoring large $\Delta n$ that carries the true information. However, excessive sparsity deteriorated the performance of forward propagation in both training and testing for an equal number of neurons because it reduces the effective capacity of the model [28]. More information was discarded in forward propagation by using high $s$ values. Therefore, there exists an optimized $s$ values in Fig. 5. Furthermore, for a high ANL, sigmoid outperformed ReLU on the classification accuracy when shifting is employed, as shown in Fig. 4 and Table 1. ReLU is widely adopted in deep neural networks because its constant derivative overcomes the vanishing gradient problem [26, 28]. However, the vanishing gradient problem is not a concern for the simple MLP demonstrated here. The major difference between ReLU and a shifted sigmoid is that ReLU is unbounded with a constant derivative for high positive values. Therefore, $\Delta n$ is not equal to zero according to Eqs. (10)-(13) even if the output neuron values exceed the highest possible target values. An upper bounded activation function with zero derivation for high positive values, such as shifted sigmoid, could suppress this unnecessary update when the nonideal training noise is significant. Our preliminary data (not shown) suggest that using a clipped ReLU activation by setting a suitable upper bound value could achieve a comparable accuracy as that using the shifted sigmoid in Fig. 4.

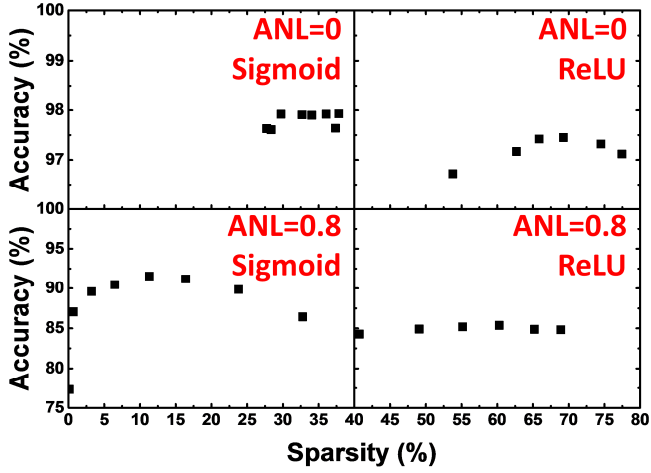

Fig. 5. Dependence of average sparsity on the classification accuracy for different activation functions and asymmetric nonlinearity.

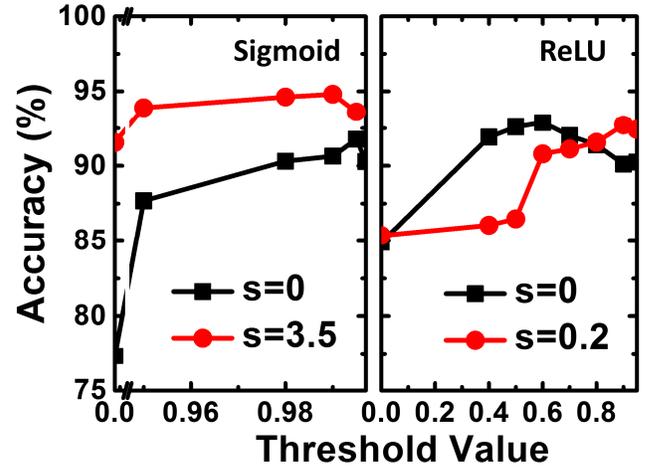

Fig. 6. Effects of threshold weight update scheme with both the sigmoid and ReLU activation functions in the presence of high asymmetric nonlinearity (ANL=0.8). The improvement of using the shifted sigmoid (with the optimized $s$=3.5) and the threshold update are cumulative while the additional benefit of using shifted ReLU (with the optimized $s$=0.2) is less pronounced.

## V. Threshold Weight Update Scheme

Inspired by the result that high sparsity is only required during weight update but not forward propagation, we propose a threshold weight update scheme where high degree of sparsity is only introduced to suppress the noise of weight update by modifying Eqs. (12)-(13) as:

$$\Delta n_{ij} = round\left[ f_{th}(X_i) \times f_{th}(H_j^{BP}) \right] \quad (15)$$

$$\Delta n_{jk} = round\left[ f_{th}(H_j) \times f_{th}(O_k^{BP}) \right] \quad (16)$$

The cutoff threshold function $f_{th}$ with a threshold value $th$ is given by:

$$\begin{aligned} f_{th}(x) &= x \quad \text{for } |x| \geq th \\ &= 0 \quad \text{for } |x| < th \end{aligned} \quad (17)$$

The optimized $th$ value is considered as an empirical parameter similar to the learning rate $\eta$ for different datasets. The same $th$ value was applied for all $X_i$, $H_j$, $H_j^{BP}$, and $O_k^{BP}$ in this work for simplicity, but potentially different $th$ values could be optimized for each term separately. Figure 6 shows the comparison of classification accuracy by using different $th$ and $s$ values. For ANL=0.8, the highest classification accuracy of 94.8 % was obtained by using $th = 0.99$ for sigmoid ($s$=3.5), which is close to the ideal 97% accuracy with ANL=0. The highest classification accuracy of 92.9 % was obtained by using $th = 0.6$ for ReLU ($s$=0). Because of the cutoff threshold, extremely high sparsity could be induced in the backpropagation path (sparsity over 97% for the sigmoid case) to suppress update noise without compromising the accuracy of forward propagation. The effects of shifted activations and threshold weight update were cumulative especially for the sigmoid activation, which without any modification produces an inherently less sparse network and is thus more prone to noise induced by erroneous weight update. Independently tuning sparsity during the forward propagation by using the shifted activation and sparsity during weight update by using the cutoff threshold function achieved higher classification accuracy than that by tuning one of them alone. Figure 7 shows the evolution of weight distributions in the first and second fully-connected layers for the optimized networks with ANL=0 (97% accuracy) and ANL=0.8 (94.8% accuracy). The final weight distributions were surprisingly different for these two cases with a comparable accuracy. This result highlights that the modified MLP algorithm, if designed properly, is self-adaptive to high asymmetric nonlinearity of weight update.

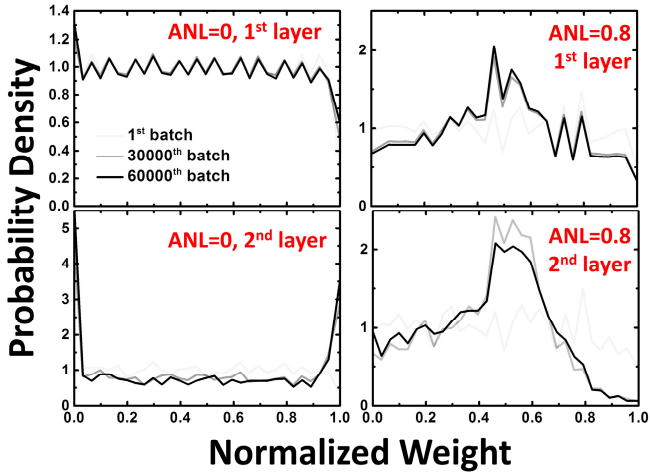

Fig. 7. Evolution of weight distributions in the first and second fully-connected layers for both ANL=0 and 0.8. The shifted sigmoid (*s*=3.5) and threshold weight update (*th*=0.99) were used.

## VI. WEIGHT AND NEURON PRECISION

The effects of reducing weight and neuron precision were further investigated in the presence of high asymmetric nonlinearity. A lower weight/neuron precision relaxes the design specifications on the synaptic devices/peripheral circuits for the ease of hardware implementation. Figure 8a shows that the requirement on the weight precision was higher for high ANL values. The asymmetric nonlinearity compromised the effective bit precision because of the irregular weight update. The future development of synaptic devices with a higher weight precision, for example by increasing the dynamic range of the conductance change to accommodate more analog states, should be continuously pursued. However, an acceptable accuracy of 87.8% was obtained by using a 6-bit weight precision, which is potentially obtainable by using practical analog RRAM-based synapses such as the one shown in Fig. 1. Figure 8b shows that the requirement on the neuron precision was also higher for high ANL values because the activations and threshold update implemented in the algorithm were sensitive to the neuron precision. A 6-bit neuron precision could be a good tradeoff between accuracy and hardware complexity.

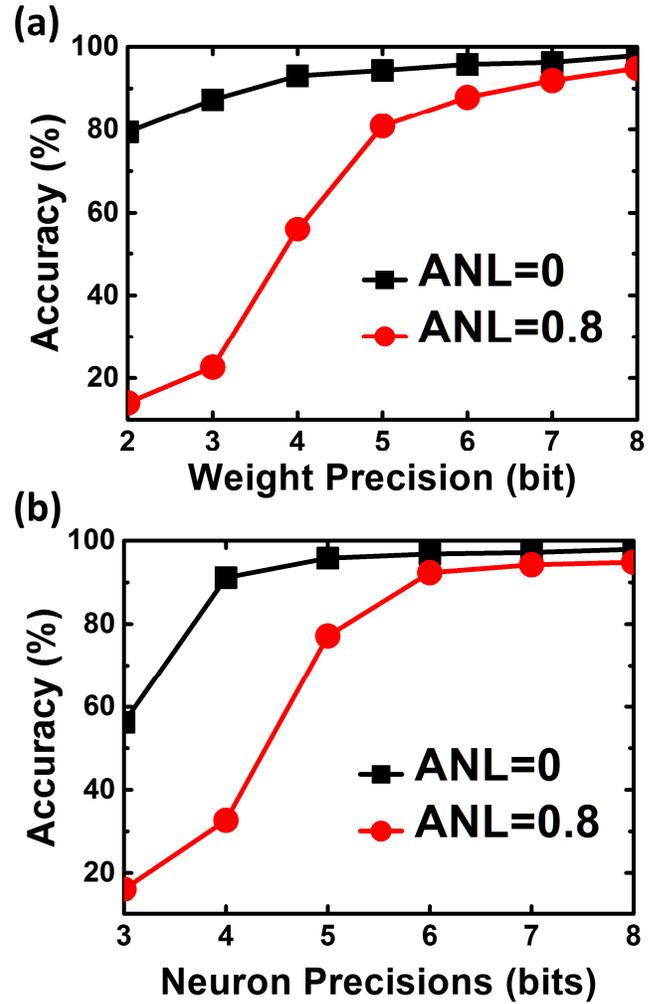

Fig. 8. Dependence of the classification accuracy on the (a) weight and (b) neuron precision. The shifted sigmoid (*s*=3.5) and threshold weight update (*th*=0.99) were used.

## VII. CONCLUSION

Asymmetric nonlinear weight update is an outstanding issue of online training for analog RRAM-based HNNs. Previous studies mainly focused on improving this non-ideal property from various perspectives of device optimization. Through careful co-optimization between device properties and hardware-applicable deep-learning algorithms, this paper proposed an alternative solution to this problem. By engineering the activation function and adopting the threshold weight update scheme, the modified MLP algorithm suppressed update noise induced by inaccurate small weight changes and is self-adaptive to asymmetric nonlinearity. Both modifications could be implemented in neuron circuits. We successfully trained a two-layer perceptron network online in the presence of extremely high asymmetric nonlinearity (ANL=0.8) and improves classification accuracy of MNIST handwritten digit dataset to 94.8% by using 8-bit weight and neuron precisions. The tradeoff between accuracy and hardware complexity was attainable using a reduced precision to 6 bits. The results suggest a promising direction for designing a robust HNN against asymmetric nonlinear weight update in practical synaptic devices.